**Exploring Multilingual Large Language Models for Enhanced TNM classification of Radiology Report in lung cancer staging**

Hidetoshi Matsuo[1], Mizuho Nishio[1,2], Takaaki Matsunaga[1], Koji Fujimoto[3], Takamichi Murakami[1]

1: Department of Radiology, Kobe University, Kobe, Japan

2: Center for Advanced Medical Engineering Research & Development, Kobe University, Kobe, Japan

3: Advanced Imaging in Medical Magnetic Resonance, Kyoto University, Kyoto, Japan

**Summary Statement**

This study investigated how language differences (English/Japanese) and the provision of TNM definitions affect TNM classification accuracy in radiology reports using zero-shot classification with GPT3.5-turbo.

**Key Results**

- This study explored the efficacy of GPT3.5-turbo, a multilingual LLM by OpenAI, in generating TNM classifications from radiology reports in both English and Japanese languages.
- Classification accuracy was significantly improved by providing the TNM definition with GPT3.5 as prompts, the classification accuracies were significantly improved.
- There was a significant decrease in N and M accuracies for Japanese reports compared with English Reports.

**Abbreviations:**

LLMs - Large Language Models

TNM - Tumor, Node, Metastasis

OR - Odds Ratio

GLMM - Generalized Linear Mixed Model

UICC - Union for International Cancer Control

**Abstract**

**Background:** Structured radiology reports remains underdeveloped due to labor-intensive structuring and narrative-style reporting. Deep learning, particularly large language models (LLMs) like GPT-3.5, offers promise in automating the structuring of radiology reports in natural languages. However, although it has been reported that LLMs are less effective in languages other than English, their radiological performance has not been extensively studied.

**Purpose**: This study aimed to investigate the accuracy of TNM classification based on radiology reports using GPT3.5-turbo (GPT3.5) and the utility of multilingual LLMs in both Japanese and English.

**Material and Methods**: Utilizing GPT3.5, we developed a system to automatically generate TNM classifications from chest CT reports for lung cancer and evaluate its performance. We statistically analyzed the impact of providing full or partial TNM definitions in both languages using a Generalized Linear Mixed Model.

**Results**: Highest accuracy was attained with full TNM definitions and radiology reports in English (M = 94%, N = 80%, T = 47%, and ALL = 36%). Providing definitions for each of the T, N, and M factors statistically improved their respective accuracies (T: odds ratio (OR) = 2.35, $p < 0.001$; N: OR = 1.94, $p < 0.01$; M: OR = 2.50, $p < 0.001$). Japanese reports exhibited decreased N and M accuracies (N accuracy: OR = 0.74 and M accuracy: OR = 0.21).

**Conclusion**: This study underscores the potential of multilingual LLMs for automatic TNM classification in radiology reports. Even without additional model training, performance improvements were evident with the provided TNM definitions, indicating LLMs' relevance in radiology contexts.

**Introduction**

Radiologists compile diagnostic reports from medical images; however, the current format of these reports, which is often narrative and unstructured, limits their utility to clinicians and patients (1). In cancer management, structuring diagnostic findings into TNM classification is crucial for clinical research and prognosis (2). Extracting TNM classifications typically demands manual effort because of the unstructured nature of radiology reports and frequent updates to classification standards, which can render previous classifications obsolete (3).

Recently, progress in deep learning in various domains has been remarkable, and its benefits have begun to be realized in real-world applications (4). Of particular note is the development after 2017, following the introduction of the transformer, which marked a significant evolution not only in natural language but also in image processing (5–9). The release of GPT-4 in 2023, which is known for its high performance, has demonstrated impressive results in various tasks, including the United States Medical Licensing Examination and the Radiology Board Examination (10,11). However, observation indicates that while this model excels in addressing questions related to clinical management, its performance notably drops in classification tasks without additional training (12).

In radiology, enhancing performance in highly domain-specific tasks, where large language models (LLMs) may lack task-relevant knowledge, can be achieved by providing additional information to the LLM (13). However, there have been few investigations into the specific knowledge that should be provided (14). Additionally, while some LLMs, like ChatGPT (15), support multiple languages, performance declines when input and output are in languages other than English, such as Japanese (11). It has been suggested that LLM performance improves with the dataset's size (16). From this perspective, it seems preferable to create models supporting multiple languages rather than specializing in a single language. However, advancing this development requires evaluating performance degradation in non-English languages using multilingual LLMs.

This study employed GPT3.5-turbo (GPT3.5), a representative multilingual LLM developed by OpenAI, to automatically classify TNM in radiology reports. We assessed GPT3.5's performance in TNM classification, considering: (i) the influence of providing either full text or parts of the TNM classification definition, and (ii) the language of the radiology reports.

**Materials and Methods**

As the evaluation of GPT3.5 utilized publicly available datasets, this study did not require ethics committee approval. We employed OpenAI's GPT3.5 to predict TNM classifications from radiology reports of chest CT scans for lung cancer in Japanese or English. We assessed the model's performance and examined how receiving TNM definitions, or portions thereof, in both languages impacted its classification ability.

**Dataset**

In this study, we utilized a dataset provided during the competition held in 2023 as part of the NII Testbeds and Community for Information Access Research (NTCIR-17), specifically for Medical Natural Language Processing for Social Media and Clinical Texts (MedNLP-SC) shared tasks. The dataset consisted of chest CT reports for lung cancer, documented in Japanese by board-certified radiologists, paired with their ground-truth TNM classifications. Although the dataset comprised 234 reports, only 162 radiology reports were accompanied by the ground truth, which was used to evaluate the zero-shot task (Figure 1).

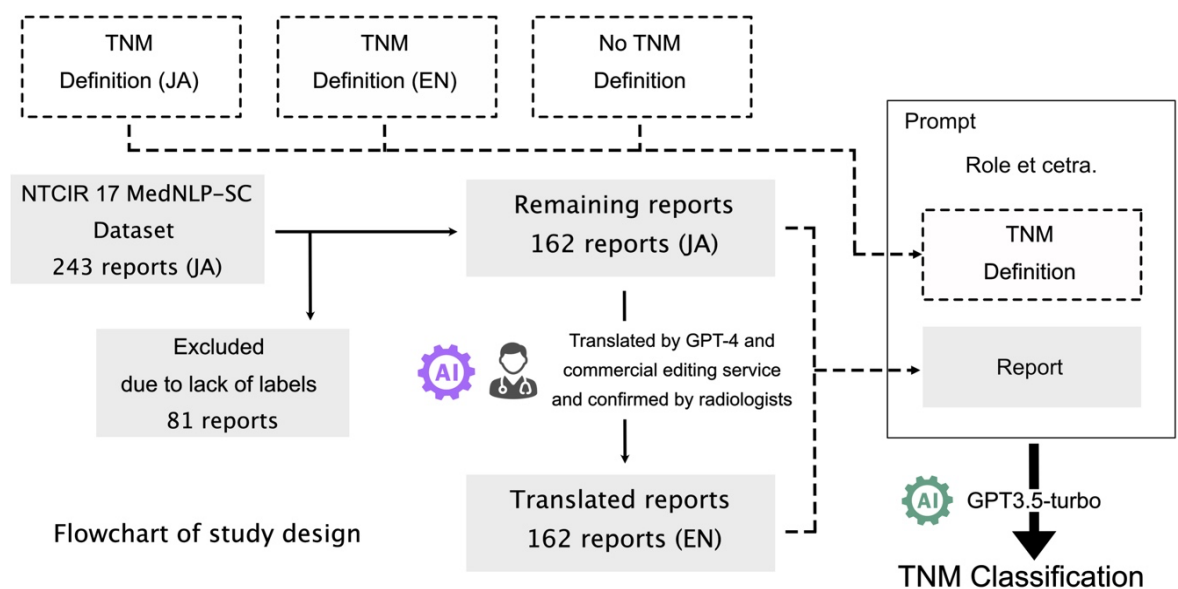

Figure 1. Flowchart of study design. NTCIR: NII Testbeds and Community for Information access Research; MedNLP-SC: Medical Natural Language Processing for Social Media and Clinical Texts; JA: Japanese language; EN: English language

TNM classifications were categorized according to the 8th edition of the UICC; however, within the MedNLP-SC dataset, subclassifications for each factor were eliminated, and T, N, and M were classified using integer values ranging from 0-4, to 0-3, and to 0-1, respectively (13).

The Japanese reports were first translated into English using OpenAI's GPT-4, linguistically reviewed by native English-speaking experts, and finally reviewed by a board-certified radiologist before being used as English Reports.

**TNM Definitions**

A simplified version of the UICC 8th edition definitions, excluding subclassifications, was created. When providing definitions, either the entire text or parts of the text were included in the prompts for GPT3.5. The definitions are as follows.

T factor (simplified version)

- T1: size of lung cancer, <3 cm
- T2: size of lung cancer, 3-5 cm
- T3: (Size of lung cancer, 5-7 cm) or (Local invasion of chest wall, parietal pericardium, phrenic nerve)
- T4: (Size of lung cancer, >7 cm) or (Invasion to the mediastinum, trachea, heart/great vessels, esophagus, vertebra, carina, recurrent laryngeal nerve)

N factor (simplified version)

- N0: no regional lymph node metastasis
- N1: metastasis in ipsilateral peribronchial and/or hilar lymph node and intrapulmonary node
- N2: metastasis in ipsilateral mediastinal and/or subcarinal lymph nodes

  N3: Metastasis in the contralateral mediastinal, contralateral hilar, ipsilateral, contralateral scalene, or supraclavicular lymph node(s)

M factor (simplified version)

- M0: no distant metastasis
- M1: distant metastasis

The TNM definitions and their Japanese translations (performed by native Japanese-speaking board-certified radiologists) were decomposed into T, N, and M factors. For each factor, the presence or absence of a definition was considered, resulting in eight combinations of TNM definitions in both Japanese (JA) and English (EN). The TNM definition sentences from the eight combinations were added to the prompts (Figure 1).

**LLM Model and Prompts**

Python with Langchain and Openai packages was used in this study. For the LLM, we employed OpenAI's GPT3.5-turbo (as of 2023/10/23). Notably, GPT3.5 demonstrates favorable results in zero-shot and few-shot learning without additional training, thus, zero-shot was employed herein. The LLM received the following prompt (only the English version is presented for brevity): The language of the prompt was the same as that of the TNM definition. Note that <Definition of TNM> was replaced with the TNM definition, and <Report> was substituted with the report (Figure 1).

Prompt (EN):

You are an experienced respiratory surgeon.

Based on the following TNM definitions, please generate a TNM classification for the given radiological report:

< Definition of TNM >

If there are no findings mentioned, assume that no abnormal findings were observed.

<Report>

Performance evaluation was conducted using all 162 reports, considering 32 different combinations that included the language of the TNM classification (EN and JA); whether each of the T, N, and M factors of the TNM classification was provided; and the language of the report documentation (EN and JA). Performance was assessed based on four criteria: whether the T, N, and M factors were correctly identified, and whether the combination of T, N, and M factors was correctly identified (ALL) (13).

**Statistical Analysis**

To statistically analyze the factors contributing to T, N, M, and ALL accuracies, a Generalized Linear Mixed Model (GLMM) was employed (17). The fixed effects included the presence or absence of T, N, and M definitions; language of the TNM definition (EN or JA); and language of the report (EN or JA), with report ID as the random effect. Statistical analyses were conducted using R (version 4.3.2) and lme4 package (version 1.1.34). Tests for T, N, M, and ALL factors applied the Bonferroni correction, with significance set at $p < 0.013$.

**Results (631/1000words)**

**Radiology Report Characteristics**

The NTCIR17 MedNLP-SC Dataset comprises 243 radiological reports. However, for this study, we required both the reports and their corresponding ground truths for TNM classification. Consequently, we excluded 81 radiology reports lacking the necessary ground truths, resulting in the utilization of 162 radiology reports from the NTCIR17 MedNLP-SC Dataset for evaluation (Figure 1). Characteristics of the radiological reports in the dataset are listed in Table 1, revealing imbalanced distributions of the T, N, and M factors. For instance, among the 162 radiology reports, the distribution of T factors was as follows: T0, 3; T1, 36; T2, 54; T3, 19; and T4, 50.

| Characteristics | Total | Ratio |
|---|---|---|
| **T factor** | | |
| 0 | 3 | 0.02 |
| 1 | 36 | 0.22 |
| 2 | 54 | 0.33 |
| 3 | 19 | 0.12 |
| 4 | 50 | 0.31 |
| **N factor** | | |
| 0 | 66 | 0.41 |
| 1 | 18 | 0.11 |
| 2 | 62 | 0.38 |
| 3 | 16 | 0.10 |
| **M factor** | | |
| 0 | 101 | 0.62 |
| 1 | 61 | 0.38 |

Table 1: Radiology Report Characteristics including T, N, and M factors.

**Performance Evaluation**

Table 2 presents the accuracies of correct TNM classifications for T, N, M, and ALL factors when the full definitions of TNM were provided, categorized by the language of the TNM definition and the radiology report. It also includes the accuracies for the T, N, M, and ALL factors when TNM definitions were not provided, categorized by the language of the radiology report. Across all the combinations, accuracies followed the order M > N > T > ALL.

When both the full TNM definition and radiology reports were in English, accuracies were as follows: M accuracy, 0.94 (153 / 162); N accuracy, 0.80 (131 / 162); T accuracy, 0.47 (76 / 162); and ALL accuracy, 0.36 (58 / 162). When both were in Japanese, the accuracies were as follows: M accuracy, 0.87 (141 / 162); N accuracy, 0.80 (129 / 162); T accuracy, 0.57 (93 / 152); and ALL accuracy, 0.43 (69 / 162). Without TNM definitions, accuracies for radiology reports in English were: M accuracy, 0.86 (139 / 162); N accuracy, 0.70 (114 / 162); T accuracy, 0.29 (47 / 162); and ALL accuracy, 0.20 (33 / 162). For radiology reports in Japanese, accuracies were: M accuracy, 0.72 (117 / 162); N accuracy, 0.57 (93 / 162); T accuracy 0.31 (50 / 162); and ALL accuracy, 0.23 (37 / 162).

These results align with the expected difficulty based on the number of available choices for each factor. Overall, providing TNM definitions led to better performance compared to when definitions were not provided (with full TNM definition: T accuracy, 0.44-0.57; N accuracy, 0.72-0.80; M accuracy, 0.85-0.95; ALL accuracy, 0.30-0.43; without TNM definition: T accuracy, 0.29-0.31; N accuracy, 0.57-0.70; M accuracy, 0.72-0.86; ALL accuracy, 0.20-0.23). Representative examples of radiology reports correctly and incorrectly predicted using GPT3.5 are shown in Figures 2 and 3, respectively. Correct predictions are evident in Figure 2(A) (T1N0M0) and Figure 2(B) (T2N1M0), while GPT3.5 incorrectly predicted all TNM factors in Figure 3(A) and inaccurately predicted the T factor in Figure 3(B).

**with full TNM definition**

| TNM definition | Report | T accuracy | N accuracy | M accuracy | ALL accuracy |
|---|---|---|---|---|---|
| EN | EN | 0.47 (76 / 162) | 0.80 (131 / 162) | 0.94 (153 / 162) | 0.36 (58 / 162) |
| EN | JA | 0.57 (92 / 162) | 0.75 (121 / 162) | 0.85 (138 / 162) | 0.39 (63 / 162) |
| JA | EN | 0.44 (71 / 162) | 0.72 (116 / 162) | 0.95 (154 / 162) | 0.30 (48 / 162) |
| JA | JA | 0.57 (93 / 162) | 0.80 (129 / 162) | 0.87 (141 / 162) | 0.43 (69 / 162) |

**without TNM definition**

| TNM definition | Report | T accuracy | N accuracy | M accuracy | ALL accuracy |
|---|---|---|---|---|---|
| - | EN | 0.29 (47 / 162) | 0.70 (114 / 162) | 0.86 (139 / 162) | 0.20 (33 / 162) |
| - | JA | 0.31 (50 / 162) | 0.57 (93 / 162) | 0.72 (117 / 162) | 0.23 (37 / 162) |

Table 2: T, N, M and ALL accuracies with and without the full TNM definitions

(A)

A lobulated tumor measuring 22 mm in its longest dimension is observed in the left upper lobe, with signs of partial pleural indentation. No lymph node enlargement is noted. No signs of pulmonary metastasis are observed. No pleural effusion is detected. No significant abnormalities are observed in the abdominal organs within the imaging range.

AI prediction: T1N0M0
Ground Truth: T1N0M0

(B)

A lobulated tumor measuring approximately 37 mm in diameter is observed in the hilum of the left upper lobe, accompanied by spiculation and retraction of the surrounding structures, which is suggestive of lung cancer. Micro-nodular opacities are observed on the dorsal side of the left lower lobe, likely old inflammatory changes caused by their relatively high density. No other obvious metastases or active lesions are present in the lung fields. The size of the lymph nodes in the left hilum is somewhat prominent, raising suspicion of metastasis. No significant mediastinal lymph node enlargement is observed. No evidence of pleural effusion exists. No obvious abnormalities are noted in the upper abdominal organs within the imaging range. No ascites is present.

A tumor of approximately 10 mm is observed in the subcutaneous area of the back, suspected to be an epidermal cyst or a similar condition.

No other significant findings are reported.

AI prediction: T2N1M0
Ground Truth: T2N1M0

Figure 2. Example of a report correctly answered by GPT3.5-turbo

(A) A 103 mm tumor is observed in the left hilar region, raising suspicion of primary lung cancer. The left main bronchus displays obstruction caused by tumor infiltration, and the left lung field is atelectatic. The tumor also appears to invade and obstruct the left pulmonary artery. Mediastinal tumor infiltration is suspected. The tumor appears to coalesce with the left hilar and left mediastinal lymph nodes. Enlargement of the right mediastinal lymph nodes also suggests possible metastasis.

Nodules scattered in the right lung field raise suspicion of secondary tumor nodules. Mild left-sided pleural effusion is detected. No significant abnormalities are observed in the upper abdominal organs within the imaging range.

AI prediction: T3N2M0
Ground Truth: T4N3M1

(B) A lobulated, irregular nodule measuring 22 mm is observed in the left upper lobe, raising suspicion of lung cancer. No other active lesions are identified in the lung fields. No significant mediastinal lymphadenopathy or pleural effusion is noted.

AI prediction: T2N0M0
Ground Truth: T1N0M0

Figure 3. Example of an incorrect answer by GPT3.5-turbo

**Statistical Analysis Using GLMM**

Tables 3 and 4 present the results of the statistical analysis conducted using the GLMM to identify factors contributing to the accuracy of T, N, M, and ALL. The odds ratios (OR) and p-values for each explanatory variable are presented in Tables 3 and 4. For T accuracy, the definition of T emerged as a significant explanatory variable that enhanced performance (OR = 2.35), whereas the definitions of N and M, languages of radiology reports, and TNM definitions were not significant explanatory variables. For N accuracy, significant improvements were observed with the definitions of N and M (OR = 1.94 and 1.27, respectively), and a significant decrease in performance was observed when reports were documented in Japanese (OR = 0.74). Regarding M accuracy, significant improvements were evident with the T and M definitions (OR = 1.52 and 2.50, respectively), particularly with the M definition. Notably, documenting reports in Japanese led to a significant decrease in performance (OR = 0.21) for M accuracy. For the overall TNM classification (ALL accuracy), the definitions of T, N, and M significantly improved performance (OR = 1.44, 1.56, and 1.28, respectively).

| Object Variable | T accuracy | |
|---|---|---|
| Explanatory Variable | Multivariable Model OR (95%CI) | P-Value |
| **TNM definition** | | |
| T | 2.35 (2.06 - 2.69) | < 0.001 |
| N | 1.16 (1.02 - 1.32) | 0.02 |
| M | 1.12 (0.98 - 1.27) | 0.09 |
| **Language** | | |
| TNM in EN | 1.00 (reference) | reference |
| TNM in JA | 1.03 (0.91 - 1.17) | 0.64 |
| Report in EN | 1.00 (reference) | reference |
| Report in JA | 1.12 (0.98 - 1.27) | 0.09 |

| Object Variable | N accuracy | |
|---|---|---|
| Explanatory Variable | Multivariable Model OR (95%CI) | P-Value |
| **TNM definition** | | |
| T | 0.97 (0.84 - 1.12) | 0.71 |
| N | 1.94 (1.68 - 2.25) | < 0.001 |
| M | 1.27 (1.10 - 1.46) | 0.001 |
| **Language** | | |
| TNM in EN | 1.00 (reference) | reference |
| TNM in JA | 0.92 (0.80 - 1.07) | 0.27 |
| Report in EN | 1.00 (reference) | reference |
| Report in JA | 0.74 (0.64 - 0.85) | < 0.001 |

| Object Variable | M accuracy | |
|---|---|---|
| Explanatory Variable | Multivariable Model OR (95%CI) | P-Value |
| **TNM definition** | | |
| T | 1.52 (1.25 - 1.85) | < 0.001 |
| N | 1.18 (0.97 - 1.43) | 0.10 |
| M | 2.50 (2.04 - 3.06) | < 0.001 |
| **Language** | | |
| TNM in EN | 1.00 (reference) | reference |
| TNM in JA | 1.12 (0.92 - 1.36) | 0.25 |
| Report in EN | 1.00 (reference) | reference |
| Report in JA | 0.21 (0.17 - 0.27) | < 0.001 |

Table 3: Statistical analysis of explanatory variables for T, N, and M accuracies using GLMM

| Object Variable | ALL accuracy | |
|---|---|---|
| Explanatory Variable | Multivariable Model OR (95%CI) | P-Value |
| **TNM definition** | | |
| T | 1.44 (1.24 - 1.68) | < 0.001 |
| N | 1.56 (1.34 - 1.82) | < 0.001 |
| M | 1.28 (1.10 - 1.49) | 0.001 |
| **Language** | | |
| TNM in EN | 1.00 (reference) | reference |
| TNM in JA | 0.94 (0.81 - 1.09) | 0.42 |
| Report in EN | 1.00 (reference) | reference |
| Report in JA | 1.11 (0.95 - 1.29) | 0.18 |

Table 4: Statistical analysis of explanatory variables for the ALL accuracy Using GLMM

## Discussion

This study showed that multilingual LLMs (GPT3.5) can achieve a significant performance in classifying TNM from radiology reports. Overall, providing TNM definitions enhances LLMs' performance in TNM classification. The improvement in LLMs performance for the T-factor is strongly influenced by the definition of T (OR 2.35), for the N-factor by the definitions of N and M (OR 1.94 and 1.27), and for the M-factor by the definitions of T and M (OR 1.52, 2.50). Consistently, definitions pertaining to the same factors lead to significant improvements. While Japanese language in radiology reports significantly decreased N and M accuracies, T and ALL accuracies were not significantly affected by language.

Previous studies reported that ChatGPT struggles, particularly in classification and calculation tasks (12). However, the results of the current study have demonstrated that providing definitions can significantly improve the accuracy of classification tasks of TNM definitions from radiology reports.

GPT3.5 demonstrated a certain level of performance even without TNM definitions, suggesting innate knowledge of TNM classification. However, as demonstrated in Table 2, supplying TNM definitions significantly enhanced GPT3.5's performance, suggesting the feasibility of imparting knowledge and definitions of TNM classification via prompts. This approach holds promise for other highly domain-specific tasks, such as TNM classification.

The LLMs performance can deteriorate when inputting and outputting languages other than English, such as Japanese (11). When assessing the impact of language on T, N, M, and ALL accuracy, a significant difference was observed in using Japanese reports for predicting N and M factors (OR = 0.74 and 0.21). In addition, there was no strong performance deterioration attributed to language in TNM definitions. Given that the radiology reports in the NTCIR17 MedNLP-SC Dataset were originally recorded in Japanese, it is speculated that the performance deterioration in this study could be linked to GPT3.5's proficiency in Japanese.

Regardless of whether TNM definitions were provided, accuracies followed the order of M > N > T > ALL, which, in conjunction with the number of available options for each factor (five for T, four for N, and two for M), suggested that T was the most challenging, followed by N and M. The definition of T contributed to the performance improvement not only in T accuracy but also in N and M accuracies, which could be partly attributed to the correlation between the characteristics of the primary tumor (T) and the patterns of its metastasis (N and M). In the present study, the definition of M contributed strongly to the accuracy of M despite its low level of difficulty. M1 was classified into three substages (M1a, M1b, and M1c) in the 8th edition of the UICC, but the M factor was simplified into two categories (M0 and M1) in this dataset. This simplification differs from the general definition of the TNM. This may be the reason for the significant improvement in the M accuracy.

Various studies have been conducted concerning the use of LLMs in the generation of radiology reports (18,19). According to the findings reported by Sun et al., when chest X-ray reports generated by GPT-4 were compared with those written by radiologists, the latter were deemed to have higher coherence, comprehensiveness, and factual consistency, as well as presented less medical harm from the radiologists' viewpoint. Conversely, referring physicians have found LLM-generated reports to be more favorable (18,19). This study can be considered an exploration of the use of LLMs to enhance the value of radiologists' reports without increasing their workload.

**Limitations**

This study had several limitations. First, the number of radiology reports (N = 162) used for the performance evaluation was small. However, this study uses zero-shot learning, which mitigates concerns regarding overfitting. Second, to observe performance fluctuations due to the English and Japanese texts, we used translated texts. While the translated texts were reviewed and corrected by specialists native to Japanese or English, the use of translated texts may affect the model's performance.

**Conclusions**

In this study, we employed GPT3.5, a representative multilingual LLM developed by OpenAI, to automatically perform TNM classification based on radiological reports. Our findings demonstrate that multilingual LLMs can achieve commendable performance in TNM classification. Even without additional training (zero-shot learning), performance enhancements were observed through strategies, such as providing TNM definitions through prompts, thereby illustrating the potential applicability of LLMs in radiology.

**Acknowledgments**

We are immensely grateful to Morteza Rohanian, Farhad Nooralahzadeh, Fabio Rinaldi, and Michael Krauthammer for their significant contributions to the conceptualization of this research. Their insights and expertise were invaluable in guiding our study.

We used OpenAI's ChatGPT (GPT-4) for translating drafts written in Japanese. OpenAI's GPT-4 was utilized for the translation of diagnostic reports, and GPT3.5-turbo was employed for TNM classification.

The present study was supported by JSPS KAKENHI (Grant Number 23K17229 and 23KK0148).